\documentclass[letterpaper]{article}
\usepackage{spconf}
\usepackage[T1]{fontenc}
\usepackage{amsmath,amssymb,graphicx,booktabs,microtype,url,array}
\usepackage{tikz,pgfplots}
\usepackage{enumitem}
\usepackage{orcidlink}
\usetikzlibrary{arrows.meta,positioning}
\pgfplotsset{compat=1.18}
\usepackage{xcolor}
\newcommand{\rev}[1]{{\color{black}#1}}

\makeatletter
\def\@ssect#1#2#3#4#5{\begingroup\bf\centering{\interlinepenalty\@M\uppercase{#5}\par}\endgroup
  \@tempskipa #3\relax\@xsect{\@tempskipa}}
\makeatother

\title{Oracle Headroom Without Signal: Null-Calibrated Evaluation of\\Candidate Selection for Thermal Heart Rate Estimation}

\name{\begin{tabular}{c} Mohammad Rakibur Rahman$^{\star}$$^{\orcidlink{0009-0003-8994-1033}}$, Nhi Nguyen$^{\star}$$^{\orcidlink{0009-0002-2090-0746}}$, Sasan Sharifipour$^{\star}$$^{\orcidlink{0000-0001-6670-4252}}$, Le Nguyen$^{\star}$$^{\orcidlink{0000-0001-7765-1483}}$,\\
Manuel Lage Cañellas$^{\star}$$^{\orcidlink{0000-0002-4917-340X}}$,  Miguel Bordallo L{\'o}pez$^{\star}$$^{\orcidlink{0000-0002-5707-9085}}$, Constantino {\'A}lvarez Casado$^{\star}$$^{\orcidlink{0000-0002-3052-4759}}$
\end{tabular}}

\address{$^{\star}$Center for Machine Vision and Signal Analysis (CMVS), University of Oulu, Finland \\
}

\begin{document}
\ninept
\maketitle
\begin{abstract}
% Overleaf abstract (235 words with comments removed, about 190 words) shortened to the 100 to 150 words of the ICASSP 2027 kit.
\rev{Camera-based physiological monitoring can produce multiple estimates from several facial regions, extraction methods, and processing settings. Signal quality indices aim to select reliable estimates without a physiological reference, and their potential is often assessed with an oracle that selects the estimate closest to the reference in each window. This retrospective selection can reward chance agreement. We model the effect with order statistics. For $K$ independent candidates unrelated to the reference, the expected oracle error decreases approximately as $1/K$. We analyze thermal heart rate estimation on 96 iBVP recordings with 168 candidates per 10\,s window. The oracle achieves a mean absolute error of 0.91\,bpm, compared with 10.74\,bpm for the best fixed configuration, 18.03\,bpm for the best quality index, and 8.61\,bpm for a constant predictor. With $K=24$, a forehead signal from another recording matches the correct one, with 4.62 against 4.61\,bpm. Oracle evaluations should report candidate count, valid coverage, and matched null controls.}

\end{abstract}


\section{Introduction}
\label{sec:introduction}
Camera-based physiological monitoring increasingly combines several extraction routes, such as facial regions, motion signals, and learned models, with several rate estimators~\cite{nguyen2025quality,elgendi2024,birrenkott2018}. No route is best in every window, so quality-aware frameworks use signal quality indices (SQIs) to decide which candidate to trust. Their evaluation often includes an oracle that uses the reference to select the most accurate candidate in each window. In our video respiratory study, this oracle reduced the mean absolute error (MAE) on the OMuSense-23 dataset from 3.03 to 0.44 \rev{breaths/min}, and the gap was attributed to imperfect quality prediction rather than to limited extractor diversity~\cite{nguyen2025quality}. \rev{Such gaps are used to justify richer quality models and larger candidate banks. An inflated gap directs effort to headroom that does not exist, and a deployed selector may then report confident estimates from signals without physiological content.}

\begin{figure}[ht!]
\centering
\input{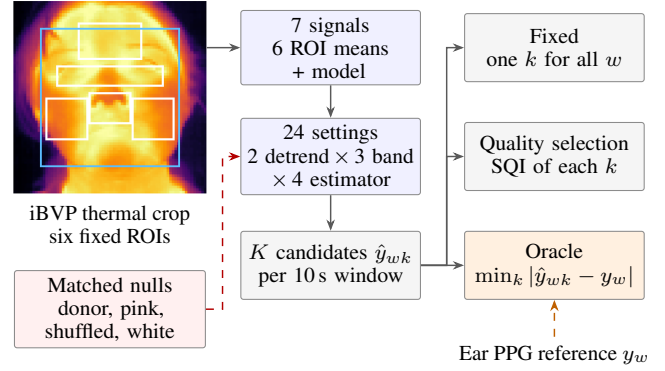}
\vspace{-6mm}
\caption{Null-calibrated oracle evaluation. A thermal crop from iBVP (recording p02a, $72\times72$ pixels) with the six fixed regions of interest (ROIs) used here, and the learned model output, give seven signals. Each passes through 24 processing settings. Fixed and quality selection use signals only. The oracle also uses the reference. Matched null inputs pass through the same settings, so that oracle errors can be compared at equal candidate count $K$.}
\label{fig:overview}
\vspace{-3mm}
\end{figure}

That interpretation assumes that the oracle measures information in the candidates. The oracle also benefits from multiplicity. When $K$ estimates lie in a plausible physiological range, one of them is likely to fall close to the reference even when none contains information on it. Related selection effects are established in model selection, circular analysis, and decoding studies that exceed theoretical chance by chance~\cite{cawley2010,kriegeskorte2009,combrisson2015}. Rate estimation is particularly exposed, because every spectral peak estimator returns a frequency inside its passband. A band-limited candidate bank is therefore plausible by construction.

Thermal facial video provides a suitable test case. The cardiac pulse has been recovered from thermal images of superficial vessels under controlled acquisition~\cite{garbey2007,chekmenev2007}, but the regional temperature averages gave weak cardiac agreement in our thermal monitoring study~\cite{casado2026} and in the associated thesis~\cite{rahman2026}. A large oracle gain on such signals should be explained before it is read as headroom.

We formulate a null-calibrated oracle evaluation and apply it to thermal heart rate estimation on iBVP~\cite{joshi2024}, as depicted in Figure~\ref{fig:overview}. The contributions are as follows:
\begin{itemize}[leftmargin=*,itemsep=0pt,topsep=1pt,parsep=0pt]
\item An order statistic model of oracle error for candidates without information about the reference, with a closed form $1/K$ approximation that gives a reference curve of uninformed candidates for any band-limited bank.
\item A measurement of oracle error against candidate count on 6,816 thermal windows with 168 candidates, compared with fixed, SQI, and constant predictors and with the model.
\item A ladder of matched nulls at the same $K$, from white noise to a thermal signal of another recording, which separates cardiac correspondence from spectral and thermal dynamics, and a reporting protocol for oracle bounds.
\end{itemize}

\section{Related Work}
\label{sec:related}
\textbf{Thermal cardiac measurement.} Thermal pulse estimation follows different physical routes. Garbey et al.~\cite{garbey2007} and Chekmenev et al.~\cite{chekmenev2007} analyzed the variation of temperature over superficial vessels. Yang et al.~\cite{yang2008} estimated a pulsatile modulation of about 0.08\,K for a camera sensitivity of 0.025\,K, and therefore selected informative pixels over the vessel and used long spectral windows. Thermal ballistocardiography instead tracks small head movements caused by cardiac ejection~\cite{pereira2018}. Regional temperature means, as used in multimodal thermal monitoring~\cite{casado2026}, average over these local effects. Other studies regress heart rate variability summaries over 1 \, min windows from facial temperature features~\cite{dicredico2022}, which can give moderate average errors without recovering a pulse waveform. iBVP provides synchronized RGB, thermal, and ear photoplethysmography (PPG) recordings to evaluate such routes~\cite{joshi2024}.

\textbf{Quality-aware selection and oracles.} SQIs describe periodicity, spectral concentration, or morphology~\cite{elgendi2024,hjorth1970}. Birrenkott et al.~\cite{birrenkott2018} learned a mapping from respiratory quality indices to estimation error, and Nguyen et al.~\cite{nguyen2025quality} extended predictive quality assessment to heterogeneous video signals and reported ground truth oracles as upper bounds. These studies compare selectors with the oracle, but do not test how much of the oracle gain remains when the candidates carry no information.

\textbf{Selection bias and permutation.} Optimising a criterion on evaluation data biases the reported score~\cite{cawley2010}. McLatchie and Vehtari~\cite{mclatchie2024} model this bias with order statistics of the candidate scores, which is the tool we apply to per window selection. Reusing the data that guided a selection distorts inference~\cite{kriegeskorte2009}. Permutation tests calibrate a score against data in which the tested correspondence is removed~\cite{ojala2010}, and empirical chance levels can exceed nominal ones in small samples~\cite{combrisson2015}. A valid permutation must rerun every step that used the data~\cite{valente2021}, and repeated measurements of one participant can inflate accuracy unless the permutation respects that nesting~\cite{jamalabadi2018}. We apply this reasoning to per window selection among physiological rate estimates, where the null must keep the candidate count and processing fixed.

\section{Null-Calibrated Oracle Evaluation}
\label{sec:method}
\subsection{Selection comparators}
Let $\hat y_{wk}$ be the heart rate given by candidate $k$ in window $w$, and $y_w$ the reference rate. For a bank $\mathcal{C}$ of $K$ candidates on $N$ windows, the best fixed and oracle MAEs are:
\begin{align}
L_{\mathrm{fix}}(\mathcal{C}) &= \min_{k\in\mathcal{C}}\frac{1}{N}\sum_{w=1}^{N}|\hat y_{wk}-y_w|,\label{eq:fixed}\\
L_{\mathrm{or}}(\mathcal{C}) &= \frac{1}{N}\sum_{w=1}^{N}\min_{k\in\mathcal{C}}|\hat y_{wk}-y_w|.\label{eq:oracle}
\end{align}
The oracle is a lower bound on the error of any selector that outputs one of the $K$ candidates. The gap $G=L_{\mathrm{fix}}-L_{\mathrm{or}}$ is commonly reported as headroom. $L_{\mathrm{fix}}$ is itself chosen on the evaluation data and is therefore a retrospective comparator.

\subsection{Oracle error without information}
Assume that in window $w$ the candidates are independent, with a distribution that does not depend on $y_w$, and let $F_w(t)=\Pr(|\hat y_{wk}-y_w|\le t)$. The minimum of $K$ such errors has expectation~\cite{david2003}:
\begin{equation}
\mathbb{E}\Big[\min_{k\le K}|\hat y_{wk}-y_w|\Big]=\int_0^{\infty}\big(1-F_w(t)\big)^K\,dt.
\label{eq:orderstat}
\end{equation}
For candidates uniform on the estimator output range $[a,b]$, and $y_w$ at distance $d_w$ from the nearest boundary, $F_w(t)=2t/(b-a)$ for $t\le d_w$. For large $K$ the integral is dominated by this interval and:
\begin{equation}
\mathbb{E}\Big[\min_{k\le K}|\hat y_{wk}-y_w|\Big]\approx\frac{b-a}{2(K+1)}.
\label{eq:uniform}
\end{equation}
With the valid range $a=45$ and $b=180$\,bpm used below, Eq.~\eqref{eq:uniform} gives 0.40\,bpm for $K=168$, and Eq.~\eqref{eq:orderstat} with $y_w=74.6$\,bpm falls below 8.61\,bpm, the constant predictor of Section~\ref{sec:results}, at $K=7$. The oracle error therefore decays as $1/K$ without any signal content, at a rate set only by the output range. Real candidates are dependent, which reduces the effective number of draws, and concentrate near typical rates, which increases $F_w$ close to $y_w$. Eq.~\eqref{eq:orderstat} is therefore a reference curve, not a bound. To summarise a measured bank, we define its uniform equivalent count $K_{\mathrm{u}}$ as the value of $K$ for which Eq.~\eqref{eq:orderstat} equals the measured $L_{\mathrm{or}}$. Redundant candidates give $K_{\mathrm{u}}<K$. Candidates concentrated near typical rates give $K_{\mathrm{u}}>K$ even without information.

\subsection{Matched nulls}
The model motivates an empirical calibration. We define the null-calibrated gain
\begin{equation}
\Delta_K=L_{\mathrm{or}}(\mathcal{C}_{\mathrm{null}})-L_{\mathrm{or}}(\mathcal{C}),
\label{eq:gain}
\end{equation}
where $\mathcal{C}_{\mathrm{null}}$ is produced by the same $K$ settings from a signal without correspondence to $y_w$, on the same windows. A positive $\Delta_K$ indicates information beyond the selection opportunity. We use a ladder of nulls that each keep more of the real signal: white noise keeps nothing, temporally shuffled samples keep the amplitude distribution, $1/f$ noise keeps the spectral shape of thermal drift, and a donor thermal signal from another recording keeps all thermal dynamics but not the temporal alignment with the reference.

\section{Experimental Setup}
\label{sec:setup}
\textbf{Data.} iBVP lists 124 paired RGB and thermal recordings~\cite{ibvp_release}. We analyze the 96 thermal recordings from 24 participants available in Rahman's cross-validation cache~\cite{rahman2026}, with four conditions each: rest with paced breathing, easy and difficult arithmetic, and guided head movement. Thermal frames have $640\times512$ pixels and 16-bit precision at approximately 30\,Hz, and the ear PPG reference is resampled to 30\,Hz~\cite{joshi2024}. Faces are cropped and resized to $72\times72$. The first 80\,s of each recording (four 600-frame clips) are divided into 10\,s windows with a 1\,s hop, giving 71 windows per recording and 6,816 windows. The reference rate is the Welch peak of the detrended PPG filtered between 0.75 and 3.00\,Hz.

\textbf{Candidate bank.} Six signals are spatial means over fixed rectangles (forehead, periorbital, perinasal, both cheeks, central face), as shown in Figure \ref{fig:overview}. The seventh is the output of EfficientPhys~\cite{efficientphys2023}, adapted to one thermal channel with rPPG-Toolbox~\cite{toolbox2023} and trained in eight folds separated by participant, with 600-frame clips, batch size of four, learning rate of 0.009, and 60 epochs~\cite{rahman2026}. Each signal is detrended with a polynomial of order one or three, filtered by a second-order Butterworth bandpass (0.75 to 3.00, 0.70 to 4.00, or 0.60 to 2.50\,Hz), and passed to four estimators: Welch~\cite{welch1967}, a windowed fast Fourier transform, multiple signal classification (MUSIC)~\cite{schmidt1986}, and the median interpeak interval. This gives $7\times2\times3\times4=168$ candidates and 1,145,088 rows. An estimate is valid when finite and between 45 and 180\,bpm. Invalid rows are excluded, so candidates can differ in coverage.

\textbf{Selectors and nulls.} The constant predictor outputs the mean reference, 74.6\,bpm. Twelve SQIs are tested as individual selectors: zero-crossing rate, Hjorth mobility and complexity~\cite{hjorth1970}, signal-to-noise ratio, band and irrelevant power ratios, periodicity, spectral entropy, peak prominence, skewness, kurtosis, and the temporal variability of the spatial 95th to 5th percentile range of the face. For the analysis of candidate count, 300 random subsets are drawn for each $K<168$, and windows with fewer than 30\% valid candidates in the full bank are removed. The four nulls of Section~\ref{sec:method} pass through the 24 settings of one signal, forming $K=24$ banks comparable with each real signal. The donor is the next recording in sorted order, which belongs to the same participant in another condition for most recordings. The code is public~\cite{oracle_code}.

\section{Results and Analysis}
\label{sec:results}
\subsection{Selection levels}
\begin{figure}[ht!]
\centering
% Fig. 3. Window rejection applied to the best fixed setting. Values from the investigation
% report (step 9, stage3_oracle.py). SQI curve uses the ranking of the best index at each level.
\begin{tikzpicture}
\begin{axis}[
 width=\columnwidth,height=4.5cm,
 xmin=48,xmax=102,ymin=0,ymax=12,x dir=reverse,
 xlabel={Windows kept (\%)},ylabel={MAE (bpm)},
 xtick={50,60,70,80,90,100},ytick={0,2,4,6,8,10,12},
 tick label style={font=\fontsize{8}{9}\selectfont},
 label style={font=\fontsize{9}{10}\selectfont},
 legend style={font=\fontsize{7.5}{8.5}\selectfont,draw=none,fill=white,fill opacity=0.85,text opacity=1,
   at={(0.02,0.04)},anchor=south west,cells={anchor=west},row sep=-1pt},
 axis line style={black!65},tick style={black!65},
 grid=major,grid style={black!9},legend cell align=left]
\addplot[black!55,densely dashed,mark=square*,mark size=1.4pt,line width=0.8pt]
 coordinates {(100,10.74) (90,10.74) (80,10.77) (70,10.72) (60,10.75) (50,10.80)};
\addlegendentry{Random rejection}
\addplot[red!75!black,mark=triangle*,mark size=1.9pt,line width=0.8pt]
 coordinates {(100,10.74) (90,10.56) (80,10.50) (70,10.59) (60,10.66) (50,10.90)};
\addlegendentry{SQI rejection}
\addplot[blue!70!black,mark=*,mark size=1.6pt,line width=0.9pt]
 coordinates {(100,10.74) (90,8.90) (80,7.51) (70,6.28) (60,5.19) (50,4.21)};
\addlegendentry{Rejection by true error}
\addplot[orange!85!black,densely dotted,line width=1pt,domain=48:102,samples=2] {8.61};
\addlegendentry{Constant predictor}
\end{axis}
\end{tikzpicture}
\vspace{-3mm}
\caption{Window rejection applied to the best fixed setting (10.74\,bpm at full coverage). Windows are discarded at random, by the SQI ranking of the best index, or by true error. Only rejection that uses the reference reduces the MAE.}
\label{fig:rejection}
\vspace{-3mm}
\end{figure}
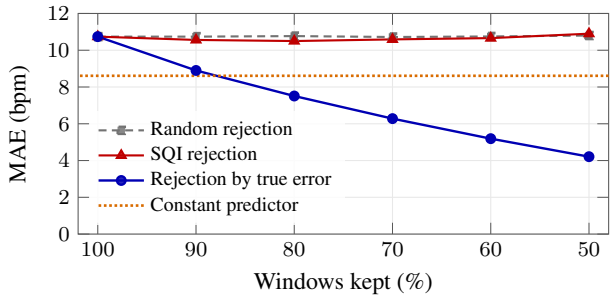

Table~\ref{tab:levels} compares the selection rules on the bank of 168 candidates. The best fixed setting (forehead, cubic detrending, 0.60 to 2.50\,Hz, MUSIC) obtains 10.74\,bpm, which is higher than the 8.61\,bpm of the constant predictor. All twelve best fixed settings use the 0.60 to 2.50\,Hz band, the narrowest tested, which limits outputs to 150\,bpm and moves them towards the population mean. The detrending order changes their MAE by 0.008 \, bpm. Individual SQI selectors obtain 18.03 to 28.30 \, bpm, so each index increases the error of the fixed setting. Figure ~\ref{fig:rejection} applies window rejection to the best fixed setting. From 100\% to 50\% coverage, SQI ranking gives 10.50 to 10.90\,bpm and random rejection 10.72 to 10.80\,bpm, while rejection by true error reaches 4.21\,bpm. The SQI curve stays within 0.25\,bpm of the random curve at every level, so the indices do not rank windows by their error. The oracle reaches 0.91\,bpm and the gap is $G=9.83$\,bpm.

\begin{table}[ht!]
\centering
\caption{Selection rules on the thermal bank of 168 candidates (6,816 windows). The last column states how the reference is used. The uniform row is the model of Eq.~\eqref{eq:orderstat}, not a measurement. Lower MAE is better.}
\vspace{1mm}
\label{tab:levels}
\setlength{\tabcolsep}{9pt}
\begin{tabular}{@{}llr@{}}
\toprule
Selection rule & Reference use & MAE (bpm)\\
\midrule
Constant, 74.6\,bpm & Mean over data & 8.61\\
Best fixed setting & One choice & 10.74\\
Best SQI (zero-crossing rate) & Index choice & 18.03\\
Oracle, $K=168$ & Every window & 0.91\\
\midrule
Uniform guesses, $K=168$ & None & 0.40\\
\bottomrule
\end{tabular}
\vspace{-3mm}
\end{table}

\subsection{Oracle error against candidate count}
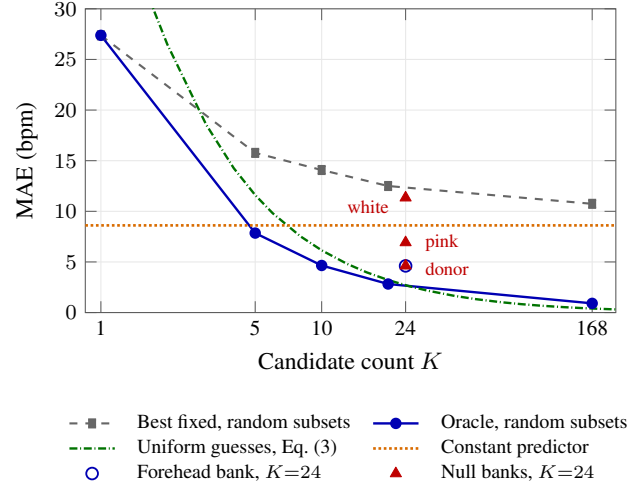
\begin{figure}[t]
\centering
% Fig. 2. Measured points are the documented summaries (README of the released code,
% investigation report steps 9 and 10). Intermediate K values are not reconstructed.
% The uninformed curve is Eq. (3) for independent candidates uniform on [45,180] bpm
% and a reference of 74.6 bpm, integrated numerically (not a fit to the data).
\begin{tikzpicture}
\begin{axis}[
 width=\columnwidth,height=5.6cm,
 xmode=log,log basis x=10,xmin=0.85,xmax=215,ymin=0,ymax=30,
 xlabel={Candidate count $K$},ylabel={MAE (bpm)},
 xtick={1,5,10,24,168},xticklabels={1,5,10,24,168},
 ytick={0,5,10,15,20,25,30},
 tick label style={font=\fontsize{8}{9}\selectfont},
 label style={font=\fontsize{9}{10}\selectfont},
 legend style={font=\fontsize{7.5}{8.5}\selectfont,draw=none,fill=none,
   at={(0.5,-0.30)},anchor=north,cells={anchor=west},row sep=-1pt,column sep=6pt},
 legend columns=2,
 axis line style={black!65},tick style={black!65},
 grid=major,grid style={black!9},
 legend cell align=left,
 every axis plot/.append style={line join=round}]
\addplot[black!60,dashed,mark=square*,mark size=1.5pt,line width=0.8pt]
 coordinates {(1,27.38) (5,15.77) (10,14.08) (20,12.50) (168,10.74)};
\addlegendentry{Best fixed, random subsets}
\addplot[blue!70!black,mark=*,mark size=1.7pt,line width=0.9pt]
 coordinates {(1,27.38) (5,7.85) (10,4.65) (20,2.82) (168,0.91)};
\addlegendentry{Oracle, random subsets}
\addplot[green!45!black,densely dashdotted,line width=0.9pt]
 coordinates {(1,44.390) (1.5,33.378) (2,26.483) (3,18.552) (4,14.253) (5,11.603) (6,9.813) (8,7.542) (10,6.147) (13,4.823) (16,3.971) (20,3.214) (25,2.596) (30,2.177) (40,1.646) (50,1.324) (65,1.023) (80,0.833) (100,0.668) (125,0.536) (150,0.447) (168,0.399) (215,0.312)};
\addlegendentry{Uniform guesses, Eq.~(3)}
\addplot[orange!85!black,densely dotted,line width=1pt,domain=0.85:215,samples=2] {8.61};
\addlegendentry{Constant predictor}
\addplot[only marks,mark=o,mark size=2.2pt,blue!70!black,line width=0.7pt]
 coordinates {(24,4.61)};
\addlegendentry{Forehead bank, $K{=}24$}
\addplot[only marks,mark=triangle*,mark size=2.4pt,red!75!black]
 coordinates {(24,4.62) (24,6.92) (24,11.34)};
\addlegendentry{Null banks, $K{=}24$}
\node[font=\fontsize{7}{8}\selectfont,red!75!black,anchor=east] at (axis cs:22,10.4) {white};
\node[font=\fontsize{7}{8}\selectfont,red!75!black,anchor=west] at (axis cs:27,6.92) {pink};
\node[font=\fontsize{7}{8}\selectfont,red!75!black,anchor=west] at (axis cs:27,4.3) {donor};
\end{axis}
\end{tikzpicture}
\vspace{-2mm}
\caption{Oracle and fixed MAE against candidate count. Filled markers are means over 300 random subsets of the bank of 168 candidates (single evaluation at $K=168$). The dash-dotted curve is Eq.~\eqref{eq:orderstat} for independent guesses uniform on 45 to 180\,bpm. Markers at $K=24$ are banks from one signal in Table~\ref{tab:controls}.}
\label{fig:candidates}
\vspace{-3mm}
\end{figure}

Figure ~\ref{fig:candidates} shows oracle and fixed errors for random subsets of the bank. The oracle MAE decreases from 27.38\,bpm at $K=1$ to 7.85, 4.65, 2.82, and 0.91\,bpm at $K=5$, 10, 20, and 168. It falls below the constant predictor at $K=5$. With ten candidates the gap is 9.43\,bpm, 96\% of the gap of the full bank, although the oracle error is still five times that of the full bank.

The measured oracle follows the uninformed curve of Eq.~\eqref{eq:orderstat} closely. The uniform equivalent counts are $K_{\mathrm{u}}=7.7$, 13.5, 22.9, and 73 for the nominal $K$ of 5, 10, 20, and 168. At small $K$ the bank behaves like slightly more than $K$ uniform guesses, which is consistent with estimates that concentrate below 150\,bpm. At large $K$ it behaves like fewer, because candidates that share a signal or a band are dependent.

\subsection{Matched nulls}
Table~\ref{tab:controls} applies the null ladder at $K=24$. White and shuffled signals reach oracle errors of 11.34 and 11.05\,bpm. Pink noise reaches 6.92\,bpm, and the donor thermal signal 4.62\,bpm. The real forehead signal obtains 4.61\,bpm, so $\Delta_{24}=0.01$\,bpm with respect to the donor. The donor also gives the lowest fixed error, 10.62\,bpm. The six real regions obtain oracle errors from 4.19 to 5.30\,bpm, all within 0.7\,bpm of the donor. The improvement from white noise to real regions is therefore reproduced by spectral shape and thermal dynamics, without temporal correspondence to the reference. The learned model gives the weakest real bank, with 9.46\,bpm and 61.6\% valid rows.

\begin{table}[t]
\centering
\caption{Matched null ladder with $K=24$ settings per signal. Fixed and oracle MAEs are in bpm. The second column lists what each signal keeps from real thermal data. Lower is better.}
\vspace{1mm}
\label{tab:controls}
\setlength{\tabcolsep}{4pt}
\begin{tabular}{@{}llrr@{}}
\toprule
Signal & Keeps & Fixed & Oracle\\
\midrule
White noise & Nothing & 17.91 & 11.34\\
Shuffled forehead & Amplitudes & 18.02 & 11.05\\
Pink noise & $1/f$ spectrum & 12.00 & 6.92\\
Donor forehead & Thermal dynamics & 10.62 & 4.62\\
\midrule
Forehead & All & 10.74 & 4.61\\
Perinasal & All & 10.99 & 4.19\\
Other four regions & All & 11.28--12.40 & 4.89--5.30\\
EfficientPhys & All & 16.52 & 9.46\\
\bottomrule
\end{tabular}
\vspace{-3mm}
\end{table}

\subsection{Thermal signal diagnostics}
Table~\ref{tab:diagnostics} summarizes the checks of the signal chain. In native 16-bit frames, the dominant thermal peak remains between 45.0 and 52.1\,bpm while the reference varies from 69.9 to 85.5\,bpm, and the thermal cardiac band holds 2 to 17\% of the power, against 86 to 97\% for the reference. Across all recordings, regional estimates are uncorrelated with the reference. Neither temporal lags of up to 30\,s nor a change of frame rate from 30 to 7.5\,Hz recovers agreement. EfficientPhys remained at a negative Pearson loss near 1.0, that is, zero correlation, in all eight folds. As a positive control, the same regional pipeline in the respiratory band gives lower error for matched recordings than for donors from other participants in 15 of 20 recordings (Wilcoxon $p=0.021$).

\begin{table}[ht!]
\centering
\caption{Diagnostics of the thermal signal chain. Protocols differ from Tables~\ref{tab:levels} and~\ref{tab:controls}. Rate errors are MAEs, and $r$ is the Pearson correlation.}
\vspace{1mm}
\label{tab:diagnostics}
\setlength{\tabcolsep}{3pt}
\renewcommand{\arraystretch}{1.05}
\begin{tabular}{@{}>{\raggedright\arraybackslash}p{3.35cm}>{\raggedright\arraybackslash}p{4.95cm}@{}}
\toprule
Check (data) & Result\\
\midrule
Training targets, 12 clips & PPG peaks at 62.4 to 92.3\,bpm, 81 to 99\% band power\\
16-bit frames, 4 rec. & Thermal peak 45.0 to 52.1\,bpm, reference 69.9 to 85.5\,bpm\\
Six ROIs, 96 rec., 20\,s & $r$ from $-0.200$ to $0.023$, MAE 21.1 to 24.2\,bpm, constant 8.51\,bpm\\
Lags of $\pm30$\,s, 96 rec. & Best $r$ 0.003 to 0.052, donor 0.013 to 0.049\\
OMIT~\cite{casado2026}, 96 rec. & 15.24\,bpm at 30\,Hz, 14.75\,bpm at 7.5\,Hz, $r\approx-0.05$\\
EfficientPhys, 8 folds & Validation loss near 1.0 from the first epoch\\
Respiration, 20 rec. & 3.21 against 4.22 breaths/min for donors, 15 of 20 lower\\
\bottomrule
\end{tabular}
\vspace{-3mm}
\end{table}

\section{Discussion}
\label{sec:discussion}
The oracle, the fixed comparator, and the null answer different questions. The oracle states whether some candidate is close to the reference. On this bank it reaches 0.91\,bpm, while no rule without the reference improves on a constant, the curve follows the uninformed model, and a donor signal matches the real one at equal $K$. The 9.83\,bpm gap is thus explained by candidate multiplicity and thermal drift within the passband, not by recoverable cardiac content. The same arithmetic applies to banks that contain physiological signal. A signal adds to the oracle only through $\Delta_K$, and multiplicity adds the rest. Oracle MAEs from 0.44 to 1.92 breaths/min in the video respiratory study~\cite{nguyen2025quality} are therefore not interpretable as attainable accuracy until $\Delta_K$ is reported. \rev{The minimum evidence that an oracle gap reflects signal is therefore a positive $\Delta_K$ at matched coverage, with a confidence interval computed over participants rather than over windows.}

The SQI results admit a consistent interpretation. Most indices reward periodicity or spectral concentration in the filtered window. A bandpass filter applied to $1/f$ thermal drift produces a narrowband oscillation near its lower cutoff, which scores as periodic while its rate stays between 45 and 55\,bpm. Selection by such indices then favours drift over the candidates close to the reference. This explanation is untested, and a learned mapping from quality to error~\cite{birrenkott2018,nguyen2025quality} may behave differently.

We recommend that oracle results state (i) the candidate count and how candidates were formed, (ii) the valid coverage of each comparator, (iii) the oracle error of a matched null bank at the same $K$ and windows, preferably with donors from other participants, and (iv) selectors, fixed settings, and constant mean and median predictors chosen on training participants only. Item (iii) requires only rerunning the existing processing on donor signals. Before any new experiment, Eq.~\eqref{eq:uniform} gives a quick check: an oracle MAE that is not clearly below $(b-a)/(2(K+1))$ for the output range of the bank is not, by itself, evidence of information in the candidates. \rev{The calibration is inexpensive, because $\Delta_K$ reuses the candidate table and needs no model training. Eq.~\eqref{eq:orderstat} also applies to any bank confined to a passband, such as RGB, radar, or WiFi rate estimators.}

The study has limitations. The donor mapping keeps the participant for most recordings. It removes temporal correspondence, but it can keep the rate level of the participant, which is the nested structure that biases accuracy in~\cite{jamalabadi2018}. Donors from other participants in the same condition, with repeated assignments that rerun the full processing~\cite{valente2021}, are needed to form a null distribution. The matched null was computed for banks of 24 settings and not for the full bank of 168 candidates. Invalid rows are excluded per candidate, so coverage is not identical across comparators, and no confidence intervals across participants are reported. The thermal findings are limited to fixed rectangles on resized crops, one adapted learned model, and one dataset. They show that the tested routes do not track heart rate, but they do not exclude local vascular information lost by averaging. An RGB positive control on the same recordings is the next step.

\section{Conclusion}
\label{sec:conclusion}
We examined oracle selection for thermal heart rate estimation with 168 candidates per window. The oracle reached 0.91\,bpm while every rule without the reference performed worse than a constant predictor, the oracle curve followed an order statistic model of guesses without information, and a donor thermal signal matched the real one at equal candidate count. Low oracle error therefore does not establish recoverable physiological information or attainable selector accuracy. Oracle bounds become interpretable when reported with the candidate count, the valid coverage, and a matched null evaluated under the same processing. \rev{Future work will apply the protocol to the iBVP RGB recordings, where cardiac information is present, and extend the matched null to the full bank with donors from other participants.}

\section*{Acknowledgment}
The research was supported \rev{by} the Research Council of Finland Smart Video Sensorization for Secure Healthcare Monitoring (SViSenS) (Grant Number: $370277$), the Interreg Aurora ResilientEdge project (Grant Number: 20373282) and Profi7 Hybrid intelligence program (352788). The authors acknowledge the use of AI assisted tools for language editing and coding support. The Elicit platform was used during the literature search as a discovery aid for candidate records. All inclusion decisions, data extraction, synthesis, interpretation and final writing decisions were made by the authors.

% Page 5 is reserved for references only. Verify after every revision.
\clearpage
% Page 5 may hold only references, funding acknowledgements, and a Compliance with Ethical Standards statement (ICASSP 2027 paper kit).
% TODO authors: verify this wording against the iBVP data licence before submission.
\section*{\rev{Compliance with Ethical Standards}}
\rev{This study was conducted retrospectively using human subject data from the iBVP dataset~\cite{joshi2024}, obtained from its authors under their data access agreement. Ethical approval was not required for this secondary analysis, as confirmed by the licence attached to the data.}

\bibliographystyle{IEEEbib}
\bibliography{references}
\end{document}